\pdfoutput=1

\documentclass[11pt]{article}

\usepackage[preprint]{acl}

\usepackage{times}
\usepackage{latexsym}
\usepackage{graphicx}
\usepackage{bbm}

\usepackage[T1]{fontenc}

\usepackage[utf8]{inputenc}

\usepackage{microtype}

\usepackage{inconsolata}

\usepackage{graphicx}

\usepackage{amsmath}
\usepackage{amssymb}

\usepackage{graphicx}

\usepackage{multirow}
\usepackage{booktabs}
\usepackage{array}

\title{HERGC: Heterogeneous Experts Representation and Generative Completion for Multimodal Knowledge Graphs}

\author{
 \textbf{Yongkang Xiao\textsuperscript{1}},
 \textbf{Rui Zhang\textsuperscript{1}}
\\
\\
 \textsuperscript{1}University of Minnesota, Minneapolis, MN, USA 
\\
 \small{
   \textbf{Correspondence:} \href{mailto:ruizhang@umn.edu}{ruizhang@umn.edu}
 }
}

\begin{document}
\maketitle
\begin{abstract}
Multimodal knowledge graphs (MMKGs) enrich traditional knowledge graphs (KGs) by incorporating diverse modalities such as images and text. multimodal knowledge graph completion (MMKGC) seeks to exploit these heterogeneous signals to infer missing facts, thereby mitigating the intrinsic incompleteness of MMKGs. Existing MMKGC methods typically leverage only the information contained in the MMKGs under the closed-world assumption and adopt discriminative training objectives, which limits their reasoning capacity during completion. Recent large language models (LLMs), empowered by massive parameter scales and pretraining on vast corpora, have demonstrated strong reasoning abilities across various tasks. However, their potential in MMKGC remains largely unexplored. To bridge this gap, we propose \textbf{HERGC}, a flexible \textbf{H}eterogeneous \textbf{E}xperts \textbf{R}epresentation and \textbf{G}enerative \textbf{C}ompletion framework for MMKGs. HERGC first deploys a Heterogeneous Experts Representation Retriever that enriches and fuses multimodal information and retrieves a compact candidate set for each incomplete triple. It then uses a Generative LLM Predictor, implemented via either in-context learning or lightweight fine-tuning, to accurately identify the correct answer from these candidates. Extensive experiments on three standard MMKG benchmarks demonstrate HERGC’s effectiveness and robustness, achieving superior performance over existing methods.
\end{abstract}

\section{Introduction}

Knowledge graphs (KGs) represent real-world facts as triples of entities and their relations, offering a structured semantic representation \citep{nickel2015review, ji2021survey}. Multimodal knowledge graphs (MMKGs) \citep{zhu2022multi,chen2024knowledge} extend traditional KGs by incorporating additional modalities such as images and text, thereby enriching the contextual information of entities and enhancing the expressiveness of the graph. Both KGs and MMKGs have been widely adopted in various AI systems, including recommender systems \citep{wang2019kgat,sun2020multi} and large language models \citep{pan2024unifying}. Moreover, they play an increasingly important role in scientific domains, supporting downstream tasks such as biomedical interaction prediction \citep{lin2020kgnn,xiao2024fuselinker}. 

\begin{figure}[htbp]
    \centering
    \includegraphics[width=1\linewidth]{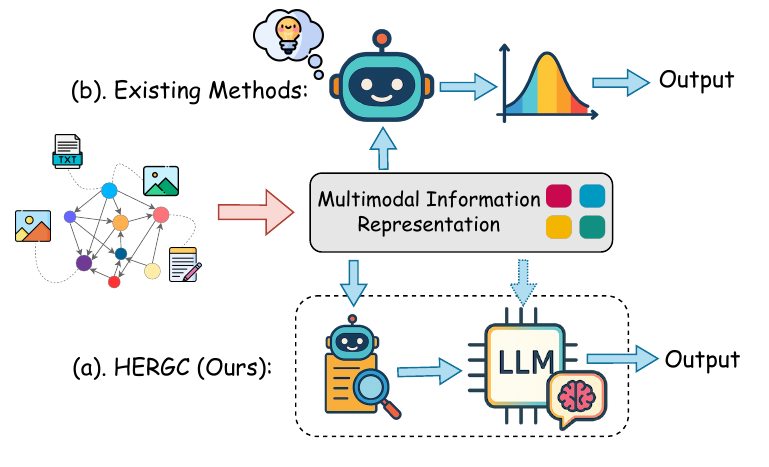} %
    \caption{Comparison between (a) HERGC and (b) existing methods. Unlike prior methods, HERGC leverages the knowledge and reasoning capabilities of LLMs to generate predictions.} 
    \label{fig:introduce} 
\end{figure}

Like traditional KGs, MMKGs also suffer from incompleteness, often due to missing facts in the underlying data sources or facts that have yet to be discovered by humans. Unlike unimodal knowledge graph completion (KGC), which primarily leverages the graph’s topological structure and local neighborhood information, multimodal knowledge graph completion (MMKGC) \citep{chen2024knowledge} introduces additional complexity through the incorporation of multimodal signals. MMKGC has advanced considerably in recent years, with most work concentrating on modality fusion \citep{li2023imf,shang2024lafa} and modality information representation \citep{zhang2025tokenization}. These approaches ultimately yield joint triple embeddings or employ ensemble strategies to score candidate facts. While these MMKGC methods offer valuable insights, their reasoning capabilities during the completion process remain limited due to their reliance solely on closed-world triples and available multimodal information.

Meanwhile, recent advances in unimodal KGC have introduced generative completion approaches powered by large language models (LLMs) \cite{wei-etal-2023-kicgpt, zhang2024making,liu2024finetuning}. By leveraging in-context learning or fine-tuning, these approaches exploit the extensive factual knowledge and reasoning capabilities that LLMs acquire during pre-training, achieving strong performance. However, this generative paradigm remains largely underexplored in the MMKGC setting due  to its inherent limitations. First, these methods typically rely on existing knowledge graph embedding (KGE) models (e.g., RotatE \cite{sun2019rotate}, as used in KICGPT \cite{wei-etal-2023-kicgpt}) for candidate retrieval, which ignore multimodal signals and yield low-recall candidate sets. Second, their LLM-based predictors are designed to process only textual inputs, excluding visual and structural modalities that are essential for comprehensive multimodal reasoning. These limitations motivate the development of novel mechanisms that seamlessly integrate multimodal information into both retrieval and generation, enabling more accurate, context‑aware reasoning within constrained candidate spaces.

To address these challenges and fill the gap in generative MMKGC, we propose HERGC, a novel and flexible generative framework that overcomes the limitations of closed‑world reasoning based solely on MMKG contents and introduces a multimodal‑aware generative completion paradigm. Figure~\ref{fig:introduce} briefly illustrates the differences between our HERGC and existing MMKGC methods.  Inspired by the retrieval-augmented generation (RAG) idea \citep{lewis2020retrieval}, HERGC comprises two core component: the Heterogeneous Experts Representation Retriever (HERR) and the Generative LLM Predictor (GLP). HERR employs a Mixture of Heterogeneous Experts Network (MoHE) to enrich each modality’s embeddings from multiple and hierarchical perspectives and a Relation-aware Gated Multimodal Unit (RaGMU) to obtain high-quality fused embeddings, which are then used to score and retrieve candidate entities. GLP supports $(1)$ directly using powerful closed-source LLMs (e.g., GPT-4) via APIs to make predictions with in-context learning, or $(2)$ injecting the fused multimodal embeddings into open-source LLMs (e.g., LLaMA) and performing LoRA fine-tuning on minimal instruction data. This flexible design allows GLP to adapt to diverse resource conditions while enabling the LLMs to accurately select the correct entity from the retrieved candidates. We conduct comprehensive experiments on three public MMKG benchmarks to validate the effectiveness and robustness of HERGC. Our contributions are summarized as follows:

\begin{itemize}
    \item We propose HERGC, the first, to the best of our knowledge, MMKGC framework based on the generative paradigm. It features a flexible Generative LLM Predictor (GLP) that supports both open-source and closed-source LLMs, enabling effective integration of external knowledge for complex multimodal structural reasoning.
    \item We design a novel Heterogeneous Experts Representation Retriever (HERR), which combines a Mixture of Heterogeneous Experts (MoHE) and a Relation-Aware Gated Multimodal Unit (RaGMU) to  extract multi-perspective signals from heterogeneous and distributionally diverse modalities, and to adaptively fuse them based on relation types, producing high-quality fused embeddings and candidate sets.
    \item We conduct extensive experiments on three standard MMKGC benchmarks, demonstrating that HERGC consistently outperforms strong baselines and exhibits robust performance across diverse settings.
\end{itemize}

\section{Related Work}
Unimodal Knowledge Graph Completion (KGC) primarily focuses on embedding entities and relations into continuous vector spaces to predict triples. Most of them leverage KG’s structure and design various score function to learn the embedding by maximize the positive and negative samples score difference, such as Translational‐Distance approaches (TransE \citep{bordes2013translating} and RotatE \citep{sun2019rotate}) and Semantic‐Matching approaches (DistMult \citep{yang2015embedding}, ComplEX \cite{trouillon2016complex}, and Tucker \cite{balavzevic2019tucker}).To improve the representation power of embedding, graph neural network based (GNN-based) methods have been proposed, such as R-GCN \citep{schlichtkrull2018modeling} and CompGCN \citep{vashishth2020composition}. 
Besides structural information, KG, as a semantic network, naturally carries text information. Therefore, the text-based method that mainly uses text information, which encodes text information in KG through a pretrained language model (PLM), has been proposed, including KG-Bert \citep{yao2019kg} and SimKGC \citep{wang2022simkgc}. With the recent development of LLM, the novel generative methods have come into view. They mainly use the rich external knowledge and powerful reasoning capabilities of LLMs to complete KGC in a sequence-to-sequence form, including KICGPT \citep{wei-etal-2023-kicgpt}, KoPA \citep{zhang2024making} and DIFT \citep{liu2019mmkg}.

\subsection{Multimodal Knowledge Graph Completion}
While recent text-based and generative approaches have started to incorporate both structural and textual information, they often lack tight coordination between these modalities during inference \citep{chen2024exploring}. Moreover,  the emergence of KGs enriched with additional modalities, such as images, audio, and video, further raises the bar for the design of dedicated MMKGC models.
Initial MMKGC models, like IKRL \cite{xie2017image}, extract visual features from entities using pre-trained visual encoders and combine these with structural embeddings. Extensions such as TransAE \citep{wang2019multimodal} and TBKGC \citep{mousselly2018multimodal} incorporate both textual and visual features, enhancing entity representations Fusion-oriented methods, including OTKGE \citep{cao2022otkge} and MoSE \cite{zhao2022mose}, employ sophisticated strategies like optimal transport and modality-specific representations to achieve effective multimodal integration. IMF \citep{li2023imf} utilizes an interactive fusion framework, training separate models for each modality to collaboratively infer missing links. Furthermore, MMKRL \citep{lu2022mmkrl} employs adversarial training but focuses specifically on robustness against modality-specific perturbations. Meanwhile, approaches like MyGO \citep{zhang2025tokenization} leverage fine-grained contrastive learning to enhance the granularity of multimodal embeddings. Also, there are methods that use multi-perspective ideas to enhance modal representation, such as MoMoK \citep{zhang2025multiple} that uses a mixture of expert model and information decoupling and MCKGC \citep{gao2025mixed} that integrates information in a mixed curvature space.

\section{Preliminary}
In this work, we focus on the most common form of Multimodal Knowledge graph (MMKG) with dual visual and textual modalities. An MMKG can be represented as a directed multigraph with modal attributes, denoted as $\mathcal{G}=(\mathcal{E},\mathcal{R},\mathcal{T},\mathcal{V},\mathcal{D})$, where $\mathcal{E}$ is the set of entities, $\mathcal{R}$ is the set of relations, and $\mathcal{T}=\{(h,r,t)|h,t \in \mathcal{E}, t \in \mathcal{R}\}$ is the set of triples (i.e. (head entity, relation, tail entity)). The $\mathcal{V}$ and $\mathcal{D}$ are the collection of visual images and descriptive text associated with entities.

Multimodal Knowledge graph Completion (MMKGC) aims to make full use of the observed triples $\mathcal{T}$ together with the visual and textual attributes of entities ($\mathcal{V}$ and $\mathcal{D}$) to infer missing triples. The set of potential facts is defined as $\{(h', r', t') \mid h', t' \in \mathcal{E},\ r' \in \mathcal{R}\}$, where $(h', r', t') \notin \mathcal{T}$ represents missing triples in the MMKG. In this work, we formulate MMKGC as the task of completing incomplete query triples of the form  $(?,r_q,t_q)$ and $(h_q,r_q,?)$, corresponding to head prediction and tail prediction, respectively. Here, we refer $h_q$ or $t_q$ the query entity and $r_q$ the the query relation.

\section{Methodology}

\begin{figure*}[htbp]
    \centering
    \includegraphics[width=1\linewidth]{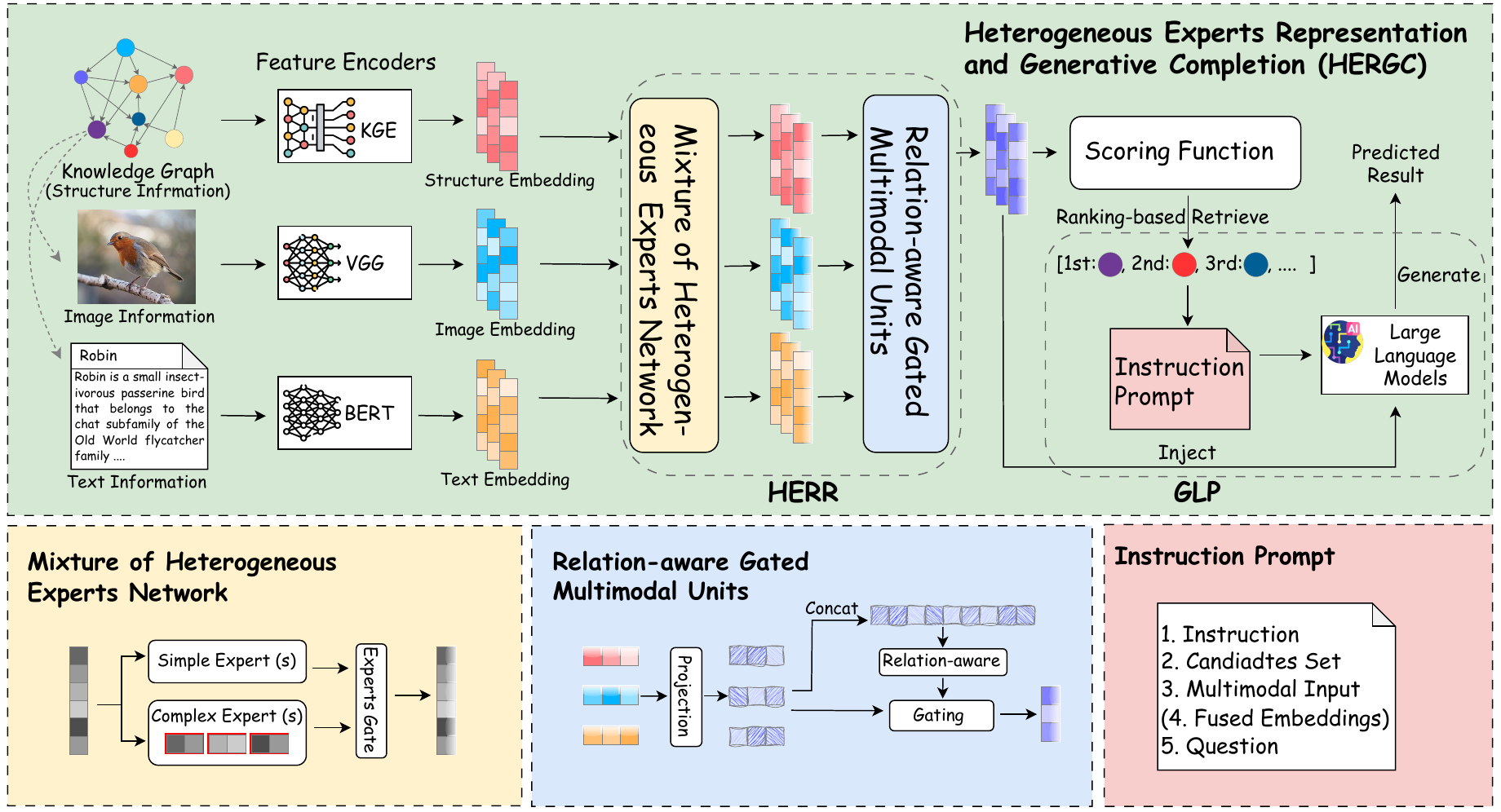} %
    \caption{Overview of the HERGC framework} 
    \label{fig:framework} 
\end{figure*}

In this section, we present the HERGC framework. We begin with the preliminary, followed by a detailed description of whole workflow. An overview of HERGC is shown in Figure~\ref{fig:framework}.


\subsection{Multimodal Information Embedding}
To enable effective fusion of multimodal information, we first encode each entity’s visual and textual modalities into embedding representations, denoted as $\mathbf{e}_\mathcal{V}$ and $\mathbf{e}_\mathcal{D}$, respectively. Additionally, we embed each entity's structural information from the KG into a structural representation $\mathbf{e}_\mathcal{S}$ to capture graph contextual cues.

\noindent \textbf{Image and Text Embedding.}
We utilize pre-trained models to encode the visual and textual information associated with each entity. To ensure a fair comparison, we maintain consistency with recent baselines \citep{zhang2025multiple,gao2025mixed} by adopting BERT \cite{devlin2019bert}, an encoder-only transformer model trained on large-scale textual corpora, for text embeddings, and VGG \citep{simonyan2014very}, a convolutional neural network trained on large-scale image datasets, for visual feature extraction. Each entity’s descriptive text and image are processed through BERT and VGG, respectively, yielding the initial modality-specific embeddings $\mathbf{e}_\mathcal{D}$ and $\mathbf{e}_\mathcal{V}$.

\noindent \textbf{Structure Embedding.}
To encode structural information from the MMKG, we adopt TuckER \citep{balavzevic2019tucker}, a representative KGE model that learns entity and relation embeddings via tensor factorization. TuckER is also employed as the scoring function in the retrieval module. The resulting embedding is taken as the structural representation $\mathbf{e}_\mathcal{S}$ for each entity.

\subsection{Heterogeneous Experts Representation Retriever}
The retriever in unimodal generative KGC typically relies on a simple KGE model to score and rank candidate triples. However, under the complex multimodal setting, maintaining high-quality entity representations and retrieval sets becomes significantly more challenging. To address this, we propose the Heterogeneous Experts Representation Retriever (HERR). HERR first enhances modality-specific features through a Mixture of Heterogeneous Experts network, then fuses them using a Relation-Aware Gated Multimodal Unit (RaGMU) to obtain joint embeddings. Finally, HERR employs a scoring function to compute triple scores and generate a ranked list of candidate entities.

\noindent \textbf{Mixture of Heterogeneous Experts Module.}
To obtain representative and multi-perspective embeddings for each modality we design a heterogeneous experts layer composed of both simple and complex experts. Given the inherent heterogeneity and distinct distributions of modality-specific embeddings, our goal is to align them across modalities while preserving their unique characteristics during fusion. To this end, we introduce the Mixture of Heterogeneous Experts (MoHE) module.

MoHE extends the standard Mixture of Experts (MoE) \citep{shazeer2017outrageously} architecture by combining diverse expert types. Given the input feature vector of the modality, MoHE outputs the weighted sum of top-$\kappa$ experts outputs:
\begin{equation}
  \label{eq:MoE1}
   \mathbf{h}_{i,m} = \sum_{\kappa \in S} G_\kappa(\mathbf{x}_{i,m})E_\kappa(\mathbf{x}_{i,m}),
\end{equation}
where $\mathbf{x}_{i,m}$ and $\mathbf{h}_{i,m}$ denote the input and output embeddings of the $i$-th entity in modality $m$, the $\mathbf{x}_{i,m}$ comes from $\mathbf{e}_{m}$, $E(\cdot)$ is the expert network, and $S=\mathrm{Top}\kappa(G_\kappa(\mathbf{x}_{i,m}))$ denotes the selected expert indices based on gating weights.

The gating weight $G_\kappa(\cdot)$ for the corresponding expert $E_\kappa$ is computed as:
\begin{equation}
  \label{eq:MoE2}
   G_\kappa(\mathbf{x}_{i,m}) = \text{softmax}(\frac{\mathbf{W}_{gate}\mathbf{x}_{i,m}+\mathbf{W}_{\epsilon}\mathbf{x}_{i,m}}{\tau}),
\end{equation}
where $\mathbf{W}_{gate}$ is the gate weight parameter matrix, $\mathbf{W}_{\epsilon}$ injects noise for exploration, and $\tau$ is the gate temperature hyperparameter.

The simple expert in the MoHE layer performs a linear transformation and feature whitening. To better adapt to heterogeneous modalities and capture richer cross-dimensional interactions, MoHE also incorporates complex PHM experts inspired by Block-Hypercomplex Linear Transformations \citep{zhangbeyond}. Specifically, the input $\mathbf{x} \in \mathbb{R}^d$ will be partitioned into $n$ sub-blocks of size $d/n$:
\begin{equation}
  \label{eq:MoE3}
   \mathbf{x} =[\mathbf{x}^{(1)};\mathbf{x}^{(2)};...;\mathbf{x}^{(n)}], \mathbf{x}^{(j)}\in \mathbb{R}^{d/n},
\end{equation}
then each sub-block is transformed by a shared weight matrix $\mathbf{W}_{block}\in \mathbb{R}^{\frac{d}{n}\times \frac{d}{n}}$ and a per-block weight matrix $\mathbf{H}_j\in \mathbb{R}^{\frac{d}{n}\times \frac{d}{n}}$:
\begin{equation}
  \label{eq:MoE4}
   \mathbf{h}^{(j)} = \mathbf{H}_j\mathbf{W}_{block}\mathbf{x}^{(j)}.
\end{equation}
Finally, the PHM expert output is obtained by concatenating all transformed sub‐blocks.
\begin{equation}
  \label{eq:MoE5}
   E_{PHM}(\mathbf{x}) = [\mathbf{h}^{(1)};\mathbf{h}^{(2)};...;\mathbf{h}^{(n)}]\in \mathbb{R}^d.
\end{equation}

\noindent \textbf{Relation-aware Gated Multimodal Units.}
In MMKGC tasks, the importance of each modality can vary across relation types. To address this, we propose the Relation-aware Gated Multimodal Unit (RaGMU), which dynamically adjusts fusion weights based on the relations.

Specifically, each modality embedding $\mathbf{x}_m$ is projected into a shared latent space:
\begin{equation}
  \label{eq:GMU1}
   \mathbf{h}_m = \tanh(\mathbf{W}_{proj,m}\mathbf{x}_m+\mathbf{b}_{proj,m})
\end{equation}
where $\mathbf{W}_{proj,m}$ and $\mathbf{b}_{proj,m}$ are the projection matrix and bias of RaGMU projector for modality $m$. 

Next, the gate vector can be calculated by:
\begin{equation}
  \label{eq:GMU2}
   \mathbf{z} = \mathrm{softmax}{(g_r(\mathbf{r}) \odot ({\mathbf{W}_{z} \mathbf{h}_{concat}+\mathbf{b}_{z}}))}
\end{equation}
where $\mathbf{h}_{concat}=[\mathbf{h}_1;\mathbf{h}_2;...;\mathbf{h}_M]$ is the concatenation of all projected modality embeddings, the $\mathbf{W}_z$ and $\mathbf{b}_z$ are the gating weight matrix and bias, $\odot$ denotes Hadamard product, and $g_r(\mathbf{r})$ is a relation-aware modulation function that generates a scaling vector from the relation embedding $\mathbf{r}$.

 Finally, the fused multimodal embedding is computed by applying the gate vector to each modality’s hidden projection $\mathbf{h}_m$:
\begin{equation}
  \label{eq:GMU3}
 \mathbf{h}_{fuse} = \sum_m\mathbf{z}_m\odot \mathbf{h}_m,
\end{equation}
where $\mathbf{z}_m$ is the $m$-th element gate vector corresponding to modality $m$.

\noindent \textbf{Score Function.}
After getting the fused multimodal embeddings, we compute triple plausibility scores using the TuckER  \citep{balavzevic2019tucker} scoring function:
\begin{equation}
  \label{eq:score1}
   S(h,r,t)=\mathbf{W}_{tucker}\times_1 \mathbf{h}_{h} \times_2 \mathbf{r} \times_3 \mathbf{h}_{t}.
\end{equation} 
where $\mathbf{h}_{h} $ and $\mathbf{h}_{t}$ denote the fused embeddings of the head $h$ and tail $t$, $\mathbf{r}_r$ denotes the embedding of the relation $r$, and $\times_n$ denotes the $n$-mode tensor product.

To train the model, we adopt a binary classification objective that encourages higher scores for positive triples and lower scores for negative ones. Negative samples are generated via uniform negative sampling. The loss function is defined as:
\begin{equation}
  \label{eq:score2}
   \mathcal{L} = -\sum[y\log\sigma(S)+(1-y)\log(1-\sigma(S))],
\end{equation}
where $y \in \{0, 1\}$ is the label indicating whether the triple is positive or negative, and $\sigma(\cdot)$ is the sigmoid function.

\subsection{Generative LLM predictor}
The Generative LLM Predictor (GLP) aims to predict the correct entity from a set of candidates given an incomplete query triple. Each query is reformulated as a natural language question derived from the query entity and relation. We adopt an instruction prompt that directly asks the LLM to complete an incomplete triple by choosing the most suitable entity. 

\noindent \textbf{Prompt Template.}
Taking the tail prediction scenario as an example, we first use HERR to retrieve the ranking of all candidates based on the query $(h_q, r_q, ?)$, ensuring that the resulting triples do not already exist in the MMKG. We then select the top-$k$ candidates, denoted as $C = [e_1, e_2, \ldots, e_k]$. A natural language question $Q$ is generated based on the query relation $r_q$ and entity $h_q$. Finally, we construct a prompt $P$ by combining the instruction $I$, the question $Q$, the candidate list $C$, and the entity descriptions $D$ (including text descriptions for $h_q$ and each $e \in C$, and the image for $h_q$ when using multimodal LLMs such as LLaMA-3-Vision):
\begin{equation}
  \label{eq:glp1}
   P=[I,Q,C,D].
\end{equation}
Using this prompt, closed-source LLMs can perform prediction via in-context learning without additional training.

\noindent \textbf{LoRA Fine-tuning.}
For open-source LLM, We perform fine‑tuning with Low‑Rank Adaptation (LoRA) on a small number of query–answer pairs. In this setting, we inject the fused embedding into the LLM via prompt using an adapter layer. Thus, the prompt template becomes:
\begin{equation}
  \label{eq:glp2}
   P=[I,Q,C,D,E],
\end{equation}
where $E$ denotes the fused embeddings of $h_q$ and each $e \in C$. This lightweight adaptation enables the model to follow our completion instruction while largely relying on its pretrained knowledge. The injected multimodal features provide additional grounding signals, guiding the model toward more accurate predictions.

\section{Experiments}

\subsection{Experiment Setup}
\textbf{Dataset.} 
We evaluate our proposed method on three benchmark MMKG datasets, MKG-Y \citep{xu2022relation}, MKG-W \citep{xu2022relation} and DB15K \citep{liu2019mmkg}. Dataset statistics and detailed descriptions are provided in Appendix~\ref{app:dataset}. 

\begin{table*}[!t]
\centering
\scriptsize
\resizebox{\textwidth}{!}{%
\begin{tabular}{ll|cccc|cccc|cccc}
\toprule
\multicolumn{2}{c}{\multirow{2}{*}{\centering \textbf{Methods}}} 
  & \multicolumn{4}{c}{\textbf{MKG-W}} 
  & \multicolumn{4}{c}{\textbf{MKG-Y}}
  & \multicolumn{4}{c}{\textbf{DB15K}} \\ 
\cmidrule(lr){3-6} \cmidrule(lr){7-10} \cmidrule(lr){11-14}
\multicolumn{2}{c}{}  
  & MRR & Hits@1 & Hits@3 & Hits@10 
  & MRR & Hits@1 & Hits@3 & Hits@10
  & MRR & Hits@1 & Hits@3 & Hits@10 \\ 
\midrule

\multicolumn{14}{c}{\textbf{Unimodal Methods}} \\ 
\cmidrule(lr){1-14}
& TransE    & 29.19 & 21.06 & 33.20 & 44.23 
            & 30.73 & 23.45 & 35.18 & 43.37
            & 24.86 & 12.78 & 31.48 & 47.07 \\ 
& RotatE    & 33.67 & 26.80 & 36.68 & 46.73 
            & 34.95 & 29.10 & 38.35 & 45.30
            & 29.28 & 17.87 & 36.12 & 49.66  \\
& DistMult  & 20.99 & 15.93 & 22.28 & 30.86 
            & 25.04 & 19.33 & 27.80 & 35.95
            & 23.03 & 14.78 & 26.28 & 39.59  \\ 
& ComplEx   & 24.93 & 19.09 & 26.69 & 36.73 
            & 28.71 & 22.26 & 32.12 & 40.93
            & 27.48 & 18.37 & 31.57 & 45.37 \\             
& TuckER    & 30.39 & 24.44 & 32.91 & 41.25 
            & 37.05 & 34.59 & 38.43 & 41.45
            & 33.86 & 25.33 & 37.91 & 50.38 \\ 

\midrule
\multicolumn{14}{c}{\textbf{Multimodal Methods}} \\ 
\cmidrule(lr){1-14}
& IKRL      & 32.36 & 26.11 & 34.75 & 44.07 
            & 33.22 & 30.37 & 34.28 & 38.26
            & 26.82 & 14.09 & 34.93 & 49.09 \\ 
& TBKGC     & 31.48 & 25.31 & 33.98 & 43.24 
            & 33.99 & 30.47 & 35.27 & 40.07
            & 28.40 & 15.61 & 37.03 & 49.86 \\ 
& TransAE   & 30.00 & 21.23 & 34.91 & 44.72
            & 28.10 & 25.31 & 29.10 & 33.03
            & 28.09 & 21.25 & 31.17 & 41.17 \\  
& MMKRL     & 30.10 & 22.16 & 34.09 & 44.69
            & 36.81 & 31.66 & 39.79 & \textbf{45.31}
            & 26.81 & 13.85 & 35.07 & 49.39 \\             
& RSME      & 29.23 & 23.36 & 31.97 & 40.43
            & 34.44 & 31.78 & 36.07 & 39.09
            & 29.76 & 24.15 & 32.12 & 40.29 \\      
& OTKGE     & 34.36 & 28.85 & 36.25 & 44.88
            & 35.51 & 31.97 & 37.18 & 41.38
            & 23.86 & 18.45 & 25.89 & 34.23 \\  
& IMF       & 34.50 & 28.77 & 36.62 & 45.44
            & 35.79 & 32.95 & 37.14 & 40.63
            & 32.25 & 24.20 & 36.00 & 48.19 \\     
& QEB       & 33.38 & 25.47 & 35.06 & 45.32
            & 34.37 & 29.49 & 37.00 & 42.30
            & 28.18 & 14.82 & 36.67 & 51.55 \\              
& VISTA     & 32.91 & 26.12 & 35.38 & 45.61
            & 30.45 & 24.87 & 32.39 & 41.53
            & 30.42 & 22.49 & 33.56 & 45.94 \\      
& MyGO      & 36.10 & 29.78 & 38.54 & 47.75
            & 38.44 & 35.01 & 39.84 & 44.19
            & 37.72 & 30.08 & 41.26 & 52.21 \\   
& MoMoK     & \underline{38.89} & 30.38 & 37.54 & 46.31
            & 37.91 & 35.09 & 39.20 & 43.20
            & 39.54 & \underline{32.38} & 43.45 & 54.14 \\   
& MCKGC     & 36.88 & 31.32 & 38.92 & 47.43
            & 38.92 & 35.49 & 40.57 & 45.21
            & 39.79 & 31.92 & 43.80 & 54.66 \\

\midrule
& HERGC\textsubscript{Retriever-only}     & 36.22 & 30.56 & 38.32 & 46.81
            & 38.42 & 35.11 & 40.16 & 44.29
            & 38.76 & 30.67 & 42.71 & 54.20   \\   
& HERGC\textsubscript{GPT-4}     & 38.28 & 32.03 & \textbf{41.80} & 47.82
            & 39.23 & 35.69 & \textbf{42.22} & 45.09
            & 39.70 & 31.22 & 45.09 & \textbf{55.38}   \\
& HERGC\textsubscript{LLaMA-3}     & \textbf{39.12} & \textbf{33.65} & \underline{41.67} & \underline{48.12}
            & \underline{39.82} & \textbf{36.73} & \underline{41.42} & 44.84
            & \textbf{40.95} & \textbf{33.47} & \textbf{45.66} & \underline{55.12}   \\ 

& HERGC\textsubscript{LLaMA-3-Vision}      & 38.76 & \underline{33.01} & 41.43  & \textbf{48.54}
            & \textbf{40.26} & \underline{36.31} & 40.91 & \underline{45.22}
            & \underline{40.28} & 32.30 & \underline{45.20} &  54.67  \\

\bottomrule
\end{tabular}
}
\caption{Main results of the comparison between HERGC and the baselines on MKG-W, MKG-Y and DB15K. For each metric, the best performance is highlighted in \textbf{bold}, and the second-best is \underline{underlined}.}
\label{tab:main_results}
\end{table*}

\noindent \textbf{Baseline Methods.}
For MMKG, we consider both the classic method based on unimodal design and the advanced method based on multimodal design. (1) For unimodal methods, we mainly consider several classic knowledge graph embedding methods: TransE \citep{bordes2013translating}, RotatE \citep{sun2019rotate}, DisMult \cite{yang2015embedding}, ComplEx \citep{trouillon2016complex} and TuckER \citep{balavzevic2019tucker}. The baseline comparisons in this paper are based on the reported performance values of these methods (2) For the multimodal methods, we selected a series of powerful multimodal KGE or KGC models: IKRL \citep{xie2017image}, TBKGC \citep{mousselly2018multimodal}, TransAE \citep{wang2019multimodal}, MMKRL \citep{lu2022mmkrl}, RSME \citep{wang2021visual}, OTKGE \citep{cao2022otkge}, IMF \citep{li2023imf}, QEB \citep{lee2023vista}, VISTA \citep{lee2023vista}, MyGO \citep{zhang2025tokenization}, MoMoK \citep{zhang2025multiple}, MCKGC \citep{gao2025mixed}. The baseline comparisons in this paper are based on the reported performance values of these methods.

\noindent \textbf{Implementation Details.}
For modality‑specific feature extraction, we use \texttt{bert‑base‑uncased} to encode text, VGG‑16 to encode images, and a TuckER model trained on the training split to obtain structural embeddings, ensuring consistency with the retriever’s scoring function. For HERR training, we tune the embedding dimension from $\{200, 300, 400\}$ and set the batch size to $\{512, 1024\}$. We use the Adam optimizer \citep{kingma2017adammethodstochasticoptimization}, with the learning rate selected from $\{0.005, 0.001, 0.0005\}$. The MoHE module is configured with 2 simple experts and 2 complex PHM experts. The number of retrieved candidate entities is selected from $\{10, 20, 30, 40\}$. For the GLP, we employ LLaMA-3-8B and apply LoRA for parameter-efficient fine-tuning. We set the LoRA hyperparameters to $r = 64$, $\alpha = 16$, a dropout rate of $0.1$, and a learning rate of $0.0002$. Additional training details are provided in Appendix~\ref{app:training}. Model performance is evaluated using standard ranking-based metrics: Mean Reciprocal Rank (MRR), and Hits@1, Hits@3, and Hits@10, under the “filtered” setting \citep{bordes2013translating}.

All experiments were conducted on an AMD EPYC 7763 64-Core CPU, an NVIDIA A100-SXM4-40GB GPU, an and Rocky Linux 8.10.

\subsection{Main Results}

Table~\ref{tab:main_results} presents the main results of our proposed HERGC compared with advanced unimodal and multimodal KGC methods. HERGC consistently achieves the best overall performance on three datasets across most evaluation metrics, demonstrating the effectiveness of its design in leveraging multimodal information and the reasoning capabilities of LLMs. Notably, HERGC improves Hits@1 on MKG-Y, MKG-W, and DB15K by $7.44\%$, $3.94\%$, and $3.37\%$, respectively, over the strongest baseline on each dataset.

We also assess the impact of different LLM predictors within GLP. LLaMA-3, after fused embeddings injection and lightweight LoRA tuning, yields consistently strong results, whereas LLaMA-3-Vision offers only marginal gains, likely because the images do not directly carry the discriminant information of the current relations. Furthermore, comparisons with GPT-4 show that with fused embeddings and lightweight fine-tuning, the open-source model can outperform the powerful closed-source model, indicating that structural reasoning ability can be enhanced through multimodal integration.

\subsection{Ablation Studies}
To verify the rationality of the HERGC design, we conduct an ablation study consisting of three parts: (1) ablation of modality-specific inputs to assess the contribution of each modality and the model’s ability to leverage multimodal information; (2) ablation of key components within HERGC, including the design of each part of the retriever and the LLM predictor; and (3) replacement of the default Tucker with alternative score functions (TransE, RotatE, and ComplEx). The results on three datasets are shown in Table~\ref{tab:combined_ablation}.

\begin{table}[ht]
\centering
\scriptsize
\resizebox{\columnwidth}{!}{%
\begin{tabular}{l|cc|cc|cc}
\toprule
\multirow{2}{*}{\textbf{Setting}} 
  & \multicolumn{2}{c|}{\textbf{MKG-W}} 
  & \multicolumn{2}{c|}{\textbf{MKG-Y}} 
  & \multicolumn{2}{c}{\textbf{DB15K}} \\ 
\cmidrule(lr){2-3} \cmidrule(lr){4-5} \cmidrule(lr){6-7}
  & MRR & Hits@1  & MRR & Hits@1  & MRR & Hits@1  \\ 
\midrule

\multicolumn{7}{c}{\textbf{Modality Information (w/o)}} \\ 
\cmidrule(lr){1-7}
Image Modality    & 36.83 & 31.19  & 38.57 & 34.96  &  39.41 & 30.74  \\
Text Modality    &  36.17 & 30.59  & 38.42 &  34.72 & 39.59  &  31.18   \\
Structure Modality    & 37.98 & 32.34  &  39.09 & 36.48 & 40.17 &32.35     \\
\midrule

\multicolumn{7}{c}{\textbf{Model Components (w/o)}} \\ 
\cmidrule(lr){1-7}
Complex Experts             & 37.95 &  32.26 & 39.04 &  35.51 & 40.26  &  32.30   \\
GMU & 37.02 &  31.41 & 38.96 & 35.19   & 39.97  &  31.28    \\
Relation-awareness   & 37.56 & 32.02 & 39.21 & 36.14 &   40.34  &  32.41   \\
Embedding Injection   & 37.94 &  32.26   & 39.00 & 35.84 & 39.05   &  31.16  \\
\midrule
\multicolumn{7}{c}{\textbf{Score Functions (w/)}} \\
\cmidrule(lr){1-7}
TransE                 & 33.27 & 26.32 & 34.78 & 28.80 & 28.58 & 20.81 \\
RotatE                 & 34.43 & 27.12 & 35.24 & 31.12 & 28.12 & 19.43 \\
ComplEx                & 27.52 & 20.03 & 31.68 & 25.78 & 32.26 & 24.50 \\

\midrule
HERGC\textsubscript{LLaMA-3}      & \textbf{39.12} & \textbf{33.65}  & \textbf{39.82} & \textbf{36.73} & \textbf{40.95} & \textbf{33.47}    \\
\bottomrule
\end{tabular}%
}
\caption{Ablation study results on three datasets, with a new group of removals above the original ones.}
\label{tab:combined_ablation}
\end{table}

For modality ablation, we individually remove the textual, visual, and structural information.  In all cases, performance declines, indicating that each modality contributes meaningfully to the model’s predictions and that HERGC effectively integrates multimodal information. For component ablation, we examine the impact of removing complex PHM experts, the RaGMU fusion module, and relation-awareness in the retriever, as well as embedding injection in the LLM predictor. Removing any of these components results in performance degradation, highlighting their importance. Notably, omitting the embedding injection also leads to a performance drop, indicating that incorporating exogenous fused multimodal embeddings enriched with graph context indeed enhances the LLM’s reasoning capability. Furthermore, the comparison of different scoring functions further validates the effectiveness of using TuckER.

\subsection{Representation Visualization}
\begin{figure}[htbp]
    \centering
    \includegraphics[width=1\linewidth]{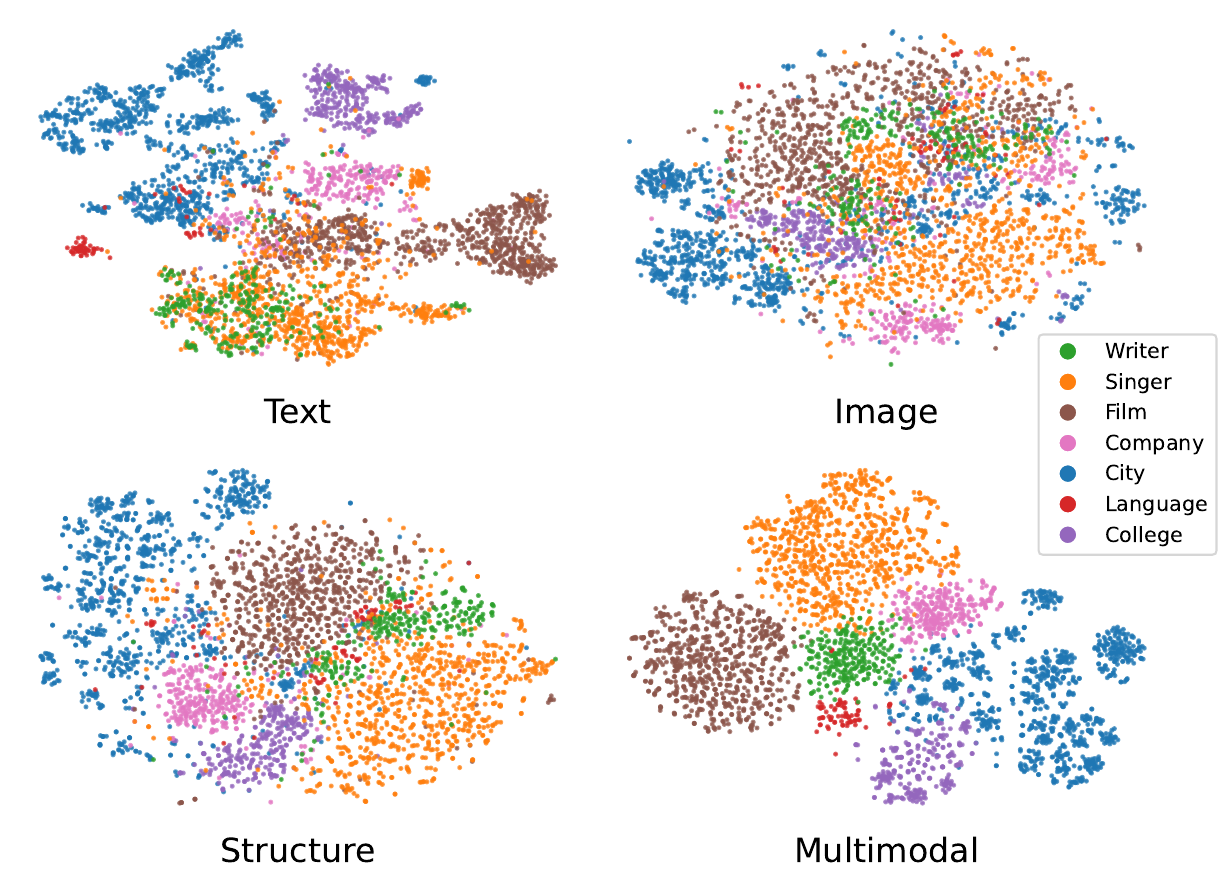} %
    \caption{t-SNE data visualization of entity representations learned by the retriever on the DB15K dataset.} 
    \label{fig:tsne} 
\end{figure}

We use t-SNE to visualize the entity representations learned by the HERR on DB15K and compared them against individual modality embeddings, providing an intuitive view to directly assess its effectiveness. We select entities from the following types: "Writer", "Singer", "Flim", "Company", "City", "Language" and "College". As shown in Figure~\ref{fig:tsne}, the fused embeddings form almost perfectly separated clusters for each entity type, with clear inter-type boundaries and uniform intra-type distributions. By contrast, image-only embeddings exhibit highly entangled regions; structure-only embeddings fail to distinguish the “Language” cluster and yield a diffuse “Writer” grouping; and text-only embeddings conflate “Writer” and “Singer” entities—likely due to their lexical similarity (e.g., names). These observations confirm that HERR effectively integrates multimodal signals to learn high-quality entity representations.

\subsection{LLM Predictor Exploration}
We further investigate the GLP when using open-source LLM by examining two factors: (1) the effect of varying the candidate set size $k$ on model performance and fine-tuning time, and (2) the impact of using LLMs with different parameter sizes.

\begin{figure}[htbp]
    \centering
    \includegraphics[width=1\linewidth]{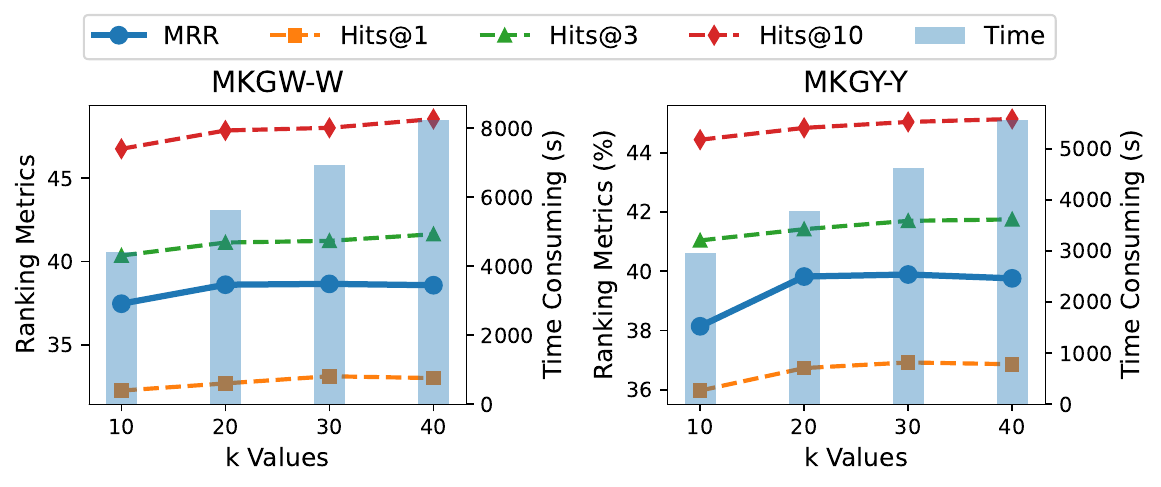} %
    \caption{The performance and time consumption of the HERGC on MKG-W and MKG-Y when $k$ takes different values.} 
    \label{fig:llmtopk} 
\end{figure}

Figure~\ref{fig:llmtopk} shows the trends in time consumption and ranking-based metrics as $k$ varies. As expected, inference time increases approximately linearly with larger $k$ values due to longer prompts constructed from larger candidate sets, which has more tokens in the prompt. However, the performance gains are marginal beyond $k=20$; only the increase from $k=10$ to $k=20$ yields a noticeable improvement in MRR. Considering the trade-off between effectiveness and efficiency, we set $k=20$ in all experiments.

 Table ~\ref{tab:llama3_3b} reports the performance and time cost of HERGC using LLMs of different scales. From the results. Although the 3 B model reduces inference time by roughly 30\% compared to the 8 B variant, it suffers a modest decline in accuracy, indicating that the more knowledge and better reasoning ability of the larger LLM is indeed helpful for MMKGC prediction.
 
\begin{table}[h]
    \centering
    \resizebox{\columnwidth}{!}{%
    \scriptsize
    \begin{tabular}{lccccc}
        \toprule
         Dataset   & MRR ($\Delta$)   & Hits@1 ($\Delta$)  & Time ($\Delta\%$) \\
        \midrule
         MKG-W    & 37.02 (-2.10)   & 31.41 (-2.24)     &  3796 (-32.8)\\   
         MKG-Y    & 38.72 (-1.10)   & 35.43 (-1.30)      & 2448 (-32.5) \\
         DB15K    & 39.61 (-1.34)   & 31.18 (-2.29)      & 5509 (-29.6)\\
        \bottomrule
    \end{tabular}%
    }
    \caption{HERGC Performance using Llama-3.2-3B as the LLM predictor ($\Delta$ values indicate differences from using Llama-3-8B).}
    \label{tab:llama3_3b}
\end{table}

\subsection{Complex Environment Simulation}
To evaluate HERGC’s robustness under realistic perturbations, we conduct complex environment simulations by: (i) injecting Gaussian noise into a fraction of the modality inputs, (ii) masking portions of the multimodal embeddings, and (iii) randomly removing a subset of training triples from the KG to emulate noisy modalities, missing multimodal information, and sparse graph connectivity, respectively.

\begin{figure}[htbp]
    \centering
    \includegraphics[width=1\linewidth]{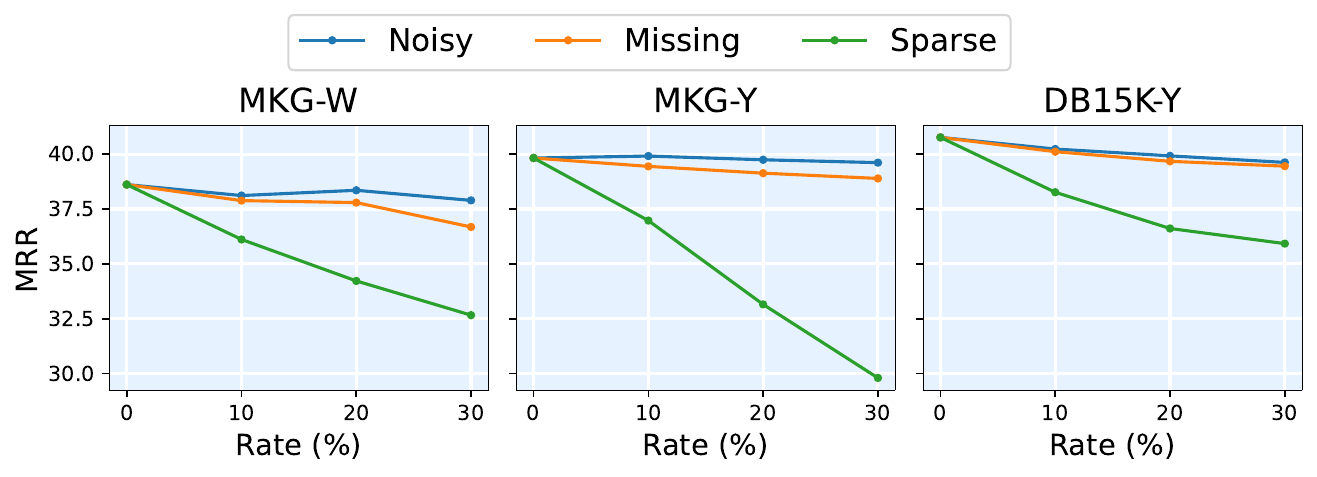} %
    \caption{Changes in MRR metrics of HERGC on three datasets under different proportions of simulated interference.} 
    \label{fig:simulation} 
\end{figure}

Figure~\ref{fig:simulation} reports how MRR degrades as we increase the proportion of corrupted modalities or removed triples. We observe that HERGC is relatively resilient to both noisy and missing multimodal inputs—its performance declines only marginally even when a substantial fraction of embeddings are perturbed or masked. In contrast, removing triples from the KG results in a visible decline in MRR, particularly on MKG-Y. When $30\%$ of the training triples are randomly removed, HERGC experiences drops of $15.4\%$, $25.1\%$, and $11.8\%$ on MKG-W, MKG-Y, and DB15K, respectively. Nevertheless, the performance degradation remains within a tolerable range, considering the inherent sensitivity of non-inductive KGC tasks to graph sparsity \citep{pujara2017sparsity}. These results highlight HERGC’s robustness and practical applicability in noisy, incomplete, and sparse multimodal scenarios.

\section{Conclusion}
In this paper, we present HERGC, a novel generative framework for MMKGC. HERGC comprises a Heterogeneous Experts Representation Retriever (HERR), which fuses multimodal signals into high-quality entity embeddings and retrieves a compact candidate set, and a Generative LLM Predictor (GLP), which predicts the correct entity from candidates and supports both open- and closed-source LLMs. Extensive experiments on three standard MMKGC benchmarks demonstrate that HERGC achieves state-of-the-art performance and consistent robustness. HERGC bridges the generative paradigm and MMKGC, providing a generalizable solution for future research.


\bibliography{acl_latex}

\newpage
\appendix

\section{Appendix}
\label{sec:appendix}

\subsection{Details of the Dataset}\label{app:dataset}

 We evaluate our proposed method on three publicly available multimodal knowledge graph completion (MMKGC) datasets: MKG-Y \citep{xu2022relation}, MKG-W \citep{xu2022relation}, and DB15K \citep{liu2019mmkg}. MKG-W (\textit{CC0 1.0 Public‑Domain Dedication)} and MKG-Y (\textit{C BY 4.0)} are curated subsets extracted from Wikidata \citep{vrandevcic2014wikidata} and YAGO \cite{suchanek2007yago} and enriched with comprehensive multimodal information including textual descriptions and associated images. DB15K (\textit{CC BY‑SA 3.0}) originates from DBpedia \citep{lehmann2015dbpedia} and similarly integrates textual and visual modalities to enhance entity representations. All three datasets provide realistic and rich multimodal scenarios, suitable for rigorous benchmarking of knowledge graph completion models. Table~\ref{tab:datasetstats} presents the statistical details of these three datasets.

\begin{table}[h]
    \centering
    \resizebox{\columnwidth}{!}{%
    \begin{tabular}{lccccc}
        \toprule
        Datasets   & Entities & Relations & Training  & Validation & Testing \\
        \midrule
        MKG-W     & 15,000   & 169        & 34,196    & 4,276      & 4,274 \\
        MKG-Y  & 15,000   & 28       & 21,310   & 2,665     & 2,663 \\
        DB15K    & 12,842    & 279        & 79,222   & 9,902     & 9,904 \\
        \bottomrule
    \end{tabular}%
    }
    \caption{Statistics of the three datasets.}
    \label{tab:datasetstats}
\end{table}

\subsection{Prompt Template}\label{app:prompt_template}
Table~\ref{tab:prompt_template} is a template with tail prediction as an example. For all three datasets, the prompt template remains consistent generally, comprising a simple instruction, a candidate set, corresponding multimodal fusion embeddings (initially represented by [Placeholder]) for reference. The only difference between the prompts for head prediction and tail prediction is that the question part is a question asking what is the head of an incomplete triple with a missing head.

\begin{table}[ht]
  \centering
  \begin{tabular}{|p{\columnwidth}|}
    \hline
    \textbf{Prompt Template for GLP} \\ \hline
    {\footnotesize\ttfamily\raggedright\sloppy
You are an excellent linguist. The task is to predict the head or tail based on the given incomplete triple, and you only need to answer one entity. The answer must be in ('candidate1', 'candidate2', 'candidate3', 'candidate4', 'candidate5', 'candidate6', 'candidate7', 'candidate8', 'candidate9', 'candidate10', 'candidate11', 'candidate12', 'candidate13', 'candidate14', 'candidate15', 'candidate16', 'candidate17', 'candidate18', 'candidate19', 'candidate20').
\\
You can refer to the entity descriptions: query entity': [image], query entity': [description], 'candidate1': [description], 'candidate2': [description], 'candidate3': [description], 'candidate4': [Placeholder], 'candidate5': [description], 'candidate6': [Placeholder], 'candidate7': [Placeholder], 'candidate8': [description], 'candidate9': [description], 'candidate10': [description], 'candidate11': [Placeholder], 'candidate12': [description], 'candidate13': [description], 'candidate14': [description], 'candidate15': [description], 'candidate16': [description], 'candidate17': [description], 'candidate18': [description], 'candidate19': [description], 'candidate20': [description]. 
\\
You can refer to the entity embeddings: 'query entity': [Placeholder], 'candidate1': [Placeholder], 'candidate2': [Placeholder], 'candidate3': [Placeholder], 'candidate4': [Placeholder], 'candidate5': [Placeholder], 'candidate6': [Placeholder], 'candidate7': [Placeholder], 'candidate8': [Placeholder], 'candidate9': [Placeholder], 'candidate10': [Placeholder], 'candidate11': [Placeholder], 'candidate12': [Placeholder], 'candidate13': [Placeholder], 'candidate14': [Placeholder], 'candidate15': [Placeholder], 'candidate16': [Placeholder], 'candidate17': [Placeholder], 'candidate18': [Placeholder], 'candidate19': [Placeholder], 'candidate20': [Placeholder]. 
\\
Question: What is the tail in ('query entity', 'query relation', tail)?
\\
Answer:
    }\\ \hline
  \end{tabular}
    \caption{Prompt template for the LLM in predictor GLP (tail prediction example).}
  \label{tab:prompt_template}
\end{table}

\subsection{Model Training}\label{app:training}
We train the retriever HERR using the training and test sets following the standard dataset splits of MKG-W, MKG-Y, and DB15K. For training the GLP when using open-source LLMs, we fine-tune the LoRA module with a small number of samples. Specifically, we employ a consistent prompt template to transform the sample triples into query–candidates formats for training. Notably, since the retriever is trained on the training set, the correct entity often receives a high score and is consistently ranked first. To prevent the LLM from overfitting to this shortcut—i.e., learning the retriever's ranking pattern rather than making predictions based on textual content—we follow previous work \citep{wei-etal-2023-kicgpt,liu2024finetuning} and use the validation set to construct the fine-tuning data for the LLM. Concretely, for MKG-W and MKG-Y, we split the original validation set into a training/validation split for LLM fine-tuning at a 9:1 ratio. For DB15K, we randomly sample 5,000 triples from its original validation set and similarly divide them into training and validation subsets using a 9:1 ratio. The test sets remain identical to the original benchmarks, and we perform both head and tail entity prediction for each test triple, in line with standard KGC evaluation protocols.

For computational efficiency, the addition of fine-tuning LLM does not introduce significant overhead. As shown in Table~\ref{tab:efficiency}, LLM fine-tuning and inference account for only 8.05\%, 7.15\%, and 8.42\% of the total training time on MKG-W, MKG-Y, and DB15K, respectively. The total training time remains reasonable for a multimodal KGC task of this scale.

\begin{table}[ht]
\centering
\scriptsize
\begin{tabular}{lccc}
\toprule
\textbf{Dataset} & \textbf{HERR} & \textbf{GLP} & \textbf{Total} \\
\midrule
MKG-W  & 17 h 53 min & 1 h 34 min & 19 h 27 min \\
MKG-Y  & 13 h 51 min & 1 h 04 min & 14 h 55 min \\
DB15K  & 23 h 33 min & 2 h 10 min & 25 h 43 min \\
\bottomrule
\end{tabular}
\caption{Training time breakdown of the HERR and GLP when using LlaMA-3.}
\label{tab:efficiency}
\end{table}

\subsection{Evaluation Metrics}\label{app:Metrics}
We employ widely‑used ranking metrics in knowledge graph completion: Mean Reciprocal Rank (MRR) and Hits@k.  

For each test query triple $(h,r,?)$ or $(?,r,t)$, the model scores every candidate entity, producing a ranked list. All metrics are reported under the \emph{filtered} setting, where corrupted triples that already exist in the KG are removed\citep{bordes2013translating}.

\noindent \textbf{Mean Reciprocal Rank (MRR).}  
Let $\operatorname{rank}_i$ denote the position of the correct entity for the $i$‑th query in the filtered list. The reciprocal rank is $1/\operatorname{rank}_i$.  
\[
\operatorname{MRR}=\frac{1}{N}\sum_{i=1}^{N}\frac{1}{\operatorname{rank}_i},
\]
where $N$ is the total number of test queries. MRR ranges from $0$ to $1$; higher values indicate better overall ranking quality.

\noindent \textbf{Hits@k.} 
Hits@k measures the proportion of queries whose correct entity appears within the top $k$ positions:
\[
\text{Hits@k} = \frac{1}{N}\sum_{i=1}^{N}\mathbbm{1}\!\bigl[\operatorname{rank}_i \le k\bigr],
\]
where $\mathbbm{1}[\cdot]$ is the indicator function. Throughout the paper we report Hits@1, Hits@3, and Hits@10, providing a fine‑grained view of top‑rank accuracy under varying tolerance levels.

\subsection{Exploring LVM and LMM as Predictors}\label{app:Llava}
To further investigate the GLP component, we experimented with replacing the LLM in GLP with a large vision model (LVM) or a large multimodal model (LMM), enabling the predictor to directly incorporate the image of the query entity in addition to textual input. The results, presented in Table~\ref{tab:llava}, show that this modification did not lead to the expected improvements. Specifically, substituting the LLM with an LVM resulted in a marked reduction in overall performance, whereas replacing it with an LMM offered no substantial benefit, yielding only marginal gains in MRR ($+1.1 \%$) and Hits@10 ($+0.8 \%$).

\begin{table}[h]
    \centering
    \resizebox{\columnwidth}{!}{%
    \begin{tabular}{lcccc}
        \toprule
         Dataset   & MRR    & Hits@1   & Hits@3   & Hits@10  \\
        \midrule
         Llama-3-8B & 39.82  & 36.73  & 41.42     & 44.84 \\   
         Llava-1.5-7B   &  27.87   &  15.79    & 38.72     & 42.68 \\
         Llama-3.2-11B-Vision   & 40.26 &  36.31  &  40.91  & 45.22 \\
        \bottomrule
    \end{tabular}%
    }
    \caption{HERGC Performance using Llava-1.5-7B and Llama-3.2-11B-Vision as the LLM predictor on MKG-Y. }
    \label{tab:llava}
\end{table}

The performance degradation observed when replacing the LLM component in GLP with LLaVA-1.5-7B may be attributed to limitations in its backbone architecture and pre-training objectives. Specifically, LLaVA-1.5-7B utilizes Llama-2-7B as its backbone, which inherently possesses weaker language modeling capabilities compared to more advanced models such as Llama-3. Moreover, LLaVA-1.5-7B is fine-tuned primarily using CLIP-based visual features and visual-language instructions, with its pre-training tasks heavily centered on image-text alignment and visual question-answering, rather than structured relational reasoning. Consequently, even after subsequent LoRA fine-tuning, the limited number of training examples might be insufficient to effectively transition the model from merely "understanding images" toward "leveraging images for relational inference in knowledge graph completion."

Similarly, the modest performance gains achieved by replacing the LLM component with Llama-3.2-11B-Vision might be due to the already mature textual reasoning capability of its underlying model, Llama-3-8B. Given the strong inherent language modeling performance of Llama-3, the additional inclusion of visual features likely provides minimal incremental benefit for relational prediction. Although large multimodal models (LMMs) generally excel at capturing visual semantics due to extensive pre-training on image-text corpora, they are not typically fine-tuned for structured relational inference tasks such as KGC. Therefore, it remains challenging for these models to accurately extract and leverage KGC-relevant relational signals from images with only a limited number of fine-tuning samples (as imposed by the LoRA rank constraints). Another potential factor is that visual information within MMKG datasets might inherently have weak correlations with the relational semantics required by the KGC task. As a result, effectively utilizing fine-grained relational clues from entity images for MMKGC remains an open and promising research direction.

\end{document}